%% file: preprint.tex
\documentclass[a4paper]{article}

\usepackage[utf8]{inputenc}
\usepackage[english]{babel}
\usepackage{amsmath}
\usepackage[nolist]{acronym}
\usepackage{multirow}
\usepackage{float}
\usepackage{graphicx}
\usepackage{natbib}

\title
{
    Synaptic Delays for Temporal Feature Detection in Dynamic Neuromorphic Processors
}

\author{Fredrik Sandin
\thanks{EISLAB, Lule{\aa} University of Technology, Sweden}
\and
Mattias Nilsson\footnotemark[1]
}

\begin{document}

    \maketitle
    
    \input{acronyms.tex}
    
    \input{Sections/0_abstract.tex}
    
    \section{Introduction}
    \label{sec:introduction}
    \input{Sections/1_introduction.tex}

    \section{Materials and Methods}
    \label{sec:materialsAndMethods}
    \input{Sections/2_methods.tex}
    
    \section{Results}
    \label{sec:results}
    \input{Sections/3_results.tex}
    
    \section{Discussion}
    \label{sec:discussion}
    \input{Sections/4_discussion.tex}

    \input{backmatter.tex}
    
    \bibliographystyle{frontiersinSCNS_ENG_HUMS}
    \bibliography{references}
    
    
    \appendix
    
    \newpage
    
    \section{DYNAP-SE Bias Parameter Values}
    \label{sec:biasValues}
    \input{Sections/A_bias_values.tex}

\end{document}

%% file: acronyms.tex
\begin{acronym}[MPC] 

    \acro{AdEx}{Adaptive Exponential Integrate-and-Fire}
    \acro{AER}{Address-Event Representation}
    \acro{ANN}{Artificial Neural Network}

    \acro{CAM}{Content-Addressable Memory}

    \acro{DPI}{Differential Pair Integrator}
    
    \acro{FDHM}{Full Duration at Half Maximum}
    \acro{FPGA}{Field-Programmable Gate Array}
    \acro{FWHM}{Full Width at Half Maximum}
    
    \acro{IPI}{Interpulse Interval}
    \acro{ISI}{Interspike Interval}
    
    \acro{PIR}{Postinhibitory Rebound}
    
    \acro{SNN}{Spiking Neural Network}
    
\end{acronym}

%% file: Sections/0_abstract.tex
\begin{abstract}
%
Spiking neural networks implemented in dynamic neuromorphic processors are well suited for spatiotemporal feature detection and learning, for example in ultra low-power embedded intelligence and deep edge applications.
Such pattern recognition networks naturally involve a combination of dynamic delay mechanisms and coincidence detection.
%
%
Inspired by an auditory feature detection circuit in crickets, featuring a delayed excitation by postinhibitory rebound, we investigate disynaptic delay elements formed by inhibitory--excitatory pairs of dynamic synapses.
We configure such disynaptic delay elements in the DYNAP-SE neuromorphic processor and characterize the distribution of delayed excitations resulting from device mismatch.
Furthermore, we present a network that mimics the auditory feature detection circuit of crickets and demonstrate how varying synapse weights, input noise and processor temperature affects the circuit. 
Interestingly, we find that the disynaptic delay elements can be configured such that the timing and magnitude of the delayed postsynaptic excitation depend mainly on the efficacy of the inhibitory and excitatory synapses, respectively.
Delay elements of this kind can be implemented in other reconfigurable dynamic neuromorphic processors and opens up for synapse level temporal feature tuning with large fan-in and flexible delays of order 10-100~ms.
%
%
\end{abstract}

%% file: Sections/1_introduction.tex

Processing of temporal patterns in signals is a central task in perception, learning and control of behaviour in both biological and artificial systems.
Using digital processors, temporal pattern recognition involves iterative processing in a network of high-frequency switching electronic circuits, which are designed to perform precise logic and arithmetic operations.
Such deterministic high-frequency circuits have high energy-dissipation density and production cost, which are not a priori necessary for reliable pattern recognition and perception in a stochastic, unreliable and continuous environment.

Unlike digital circuits, neurons are unreliable, stochastic and slow 
information processing entities which form networks that function reliably through distributed information processing and adaptation.
%
Neural circuits are therefore interesting models for implementation in nano-electronic substrates that are subject to device mismatch and failure.
Highly energy-efficient neuromorphic processors and sensor systems are designed by matching the device dynamics to neural dynamics, for example in the form of CMOS analog circuits operating in the subthreshold regime where semiconductor electron diffusion mimics ion diffusion in biological ion channels \citep{schuman2017survey,indiveri2011neuromorphic,mead1990neuromorphic}.

The dynamic nature and spatial structure of biological neurons (synapses, dendrites, axons, etc.) implies that \acp{SNN} are inherently capable of temporal pattern recognition \citep{mauk2004temporal} and pattern generation, also without recurrent connections.
Thus, \acp{SNN} with biologically plausible dynamics offer an interesting alternative model for temporal and spatial (spatiotemporal) pattern recognition, which is compatible with neural circuits in biology.
However, it is an open question how neuromorphic engineers \citep{indiveri2011frontiers} should design and implement such SNN-based pattern recognition solutions using neuromorphic processors in practical applications.

Here we present and characterize a temporal feature detection circuit implemented in the ultra low-power Dynamic Neuromorphic Asynchronous Processor (DYNAP) model SE from aiCTX \citep{moradi2018dynaps}.
The DYNAP-SE has reconfigurable mixed-mode analog/digital neuron and synapse circuits featuring biologically faithful dynamics.
The proposed circuit mimics an auditory feature detection circuit in crickets \citep{schoneich2015auditory}, which enables reliable detection of temporal patterns of $10$--$100$~ms duration using three spiking neurons with reconfigurable synaptic delay and coincidence detection dynamics.

Temporal delays are essential for neuromorphic processing of temporal patterns in spike trains \citep{sheik2013spatio}
and have been studied since the early 90s, see for example the work by \cite{spiegel1994large}.
Temporal delays have been implemented in neuromorphic processors in the form of dedicated, specifically tuned delay neurons in the network architecture \citep{sheik2012emergent, sheik2012exploiting, coath2014robust}.
The resulting \ac{SNN} is similar to a model of the auditory thalamocortical system described by \cite{coath2011emergent}.
\cite{nielsen2017compact} present a low-power pulse delay and extension circuit for neuromorphic processors,
which implements programmable axonal delays ranging from fractions of microseconds up to tens of milliseconds.
Architectures in which asynchronously firing neurons project to a common target along delay lines so that spikes arrive at the target neuron simultaneously, and thus causing it to fire, are called polychronous \citep{izhikevich2006polychronization}.
A polychronous \ac{SNN} with delay adaptation has been implemented for spatiotemporal pattern recognition purposes in an \ac{FPGA} and in a custom mixed-signal neuromorphic processor \citep{wang2013fpga,wang2014mixed}.

Temporal processing is partially realised by delay lines between neurons.
However, the dynamics of synapses (and dendrites) also play an important role for the processing of temporal and spatiotemporal patterns \citep{mauk2004temporal} and offer efficient dynamic mechanisms for sequence detection and learning \citep{Buonomano1129}.
%
Synaptic dynamics also enables pattern recognition architectures with high fan-in/out, which is beneficial in neuromorphic systems where axon/neuron reservation and spike transmission is costly.
\cite{rost2013neuromorphic} present an \ac{SNN} architecture with spike frequency adaptation and synaptic short term plasticity that models auditory pattern recognition in cricket phonotaxis.
There, synaptic short-term depression and potentiation is implemented to make neurons act as high-pass and low-pass filters, respectively.
The resulting signals are combined in a neuron that acts as a band-pass filter and thereby responds to a frequency band that is matched to the particular sound pulse period of the crickets.
Insects offer interesting opportunities to develop neuromorphic systems by modelling and finding inspiration from their neural circuits, where the relatively low level of complexity allows neuromorphic engineers to transfer the principles of neural computation to applications \citep{dalgaty2018insect}.

Our present investigation is guided by a more recent description of the cricket auditory system \citep{schoneich2015auditory} and preliminary work \citep{nilsson2018monte}, indicating that the synaptic dynamics of the DYNAP-SE can be used to approximate the excitatory rebound dynamics of a non-spiking delay neuron in the auditory circuit of the cricket.
%
The flexible and easily configurable properties of the synaptic delay elements described and characterized in the following opens up for further development of \ac{SNN} pattern recognition architectures for neuromorphic processors.

%% file: Sections/2_methods.tex
\subsection{The DYNAP-SE Neuromorphic Processor}

The DYNAP-SE neuromorphic processor uses a combination of low-power, inhomogeneous sub-threshold analog circuits and fast, programmable digital circuits for the emulation of \ac{SNN} architectures with bio-physically realistic neuronal and synaptic behaviors \citep{moradi2018dynaps}, making it a platform for spike-based neural processing with colocalized memory and computation \citep{indiveri2015pieee}. Specifically, the DYNAP-SE comprises four four-core neuromorphic chips, each with 1k analog silicon neuron circuits. Each neuron has a \ac{CAM} block containing 64 addresses representing the presynaptic neurons that the neuron is connected to. Information about spike-activity is transmitted between neurons in an \ac{AER} digital routing scheme. Four different types of synaptic behavior are available for each connection:
fast excitatory, slow excitatory, subtractive inhibitory, and shunting inhibitory. The dynamic behaviors of the neuronal and synaptic circuits of the DYNAP-SE are governed by analog circuit parameters which are set by programmable on-chip temperature compensated bias-generators \citep{delbruck2010bias}.

The inhomogeneity of the analog low-power circuits that constitute the neurons and synapses of the DYNAP-SE neuromorphic processor is due to device mismatch, and gives rise to variations in the dynamic behaviors of the silicon neurons and synapses that the analog circuits constitute. These variations are analogous to differences in values of the parameters governing the differential equations that model the neuronal and synaptic dynamics implemented in the chips. Consequently, one set value of a neuronal or synaptic bias parameter, in one core of the DYNAP-SE, results in a distribution of the corresponding parameter values in the population of neurons and synapses of that core.

\subsubsection{Spiking Neuron Model}
In the DYNAP-SE, neurons are implemented according to the \ac{AdEx} spiking neuron model \citep{brette2005adaptive}.
The model describes the neuron membrane potential, $V$, and the adaptation variable, $w$, with two coupled nonlinear differential equations,

\begin{subequations}
	\begin{equation}
		C\frac{dV}{dt}=-g_{L}(V-E_{L})+g_{L}\Delta_{T}e^{\left( V-V_{T}\right)/\Delta_{T}}-w+I,
		\label{eq:AdEx_membrPot}
	\end{equation}
	\begin{equation}
		\tau_{w}\frac{dw}{dt}=a\left(V-E_{L}\right)-w,
		\label{eq:AdEx_adapt}
	\end{equation}
	\label{eq:AdEx}
\end{subequations}
where 
$C$ is the membrane capacitance,
$g_L$ the leak conductance,
$E_L$ the leak reversal potential,
$V_T$ the spike threshold,
$\Delta_T$ the slope factor,
$I$ the (postsynaptic) input current,
$\tau_w$ the adaptation time constant, and
$a$ the subthreshold adaptation.
The membrane potential increases rapidly for $V>V_T$ due to the nonlinear exponential term,
which leads to rapid depolarisation and spike generation at time $t = t_{spike}$,
where the membrane potential and adaptation variable are updated according to 
\begin{subequations}
	\begin{equation}
		V \rightarrow V_{r},
		\label{eq:AdEx_voltRes}	
	\end{equation}
	\begin{equation}
		w \rightarrow w+b,
		\label{eq: spike-triggered adaptation}	
	\end{equation}
	\label{eq:AdEx_spike}
\end{subequations}
where $V_r$ is the reset potential and
$b$ is the spike-triggered adaptation.

\subsubsection{Dynamic Synapse Model}

In the DYNAP-SE, synapses are implemented with sub-threshold \ac{DPI} log-domain filters described by \cite{chicca2014neuromorphic}.
The response of the \ac{DPI} for an input current $I_{in}$ can be approximated with a first-order linear differential equation,
\begin{equation}
	\tau\frac{d}{dt} I_{out} + I_{out} = \frac{I_{th}}{I_{\tau}}I_{in},
    \label{eq:DPI}
\end{equation}
where 
$I_{out}$ is the (postsynaptic) output current,
$\tau$ and 
$I_{\tau}$ are time constant parameters, and
$I_{th}$ is an additional control parameter that can be used to change the gain of the filter.
This approximation is valid in the domain where $I_{in} \gg I_{\tau}$ and $I_{out} \gg I_{I_{th}}$.
The \ac{AdEx} neuron model and the synapse equation are used in the following to describe the synaptic delay elements that we configure in the DYNAP-SE in order to approximate the cricket auditory feature detection circuit.



\subsection{Cricket Auditory Feature Detection Circuit}

\begin{figure}[h!]
    \begin{center}
        \includegraphics[width=\textwidth]{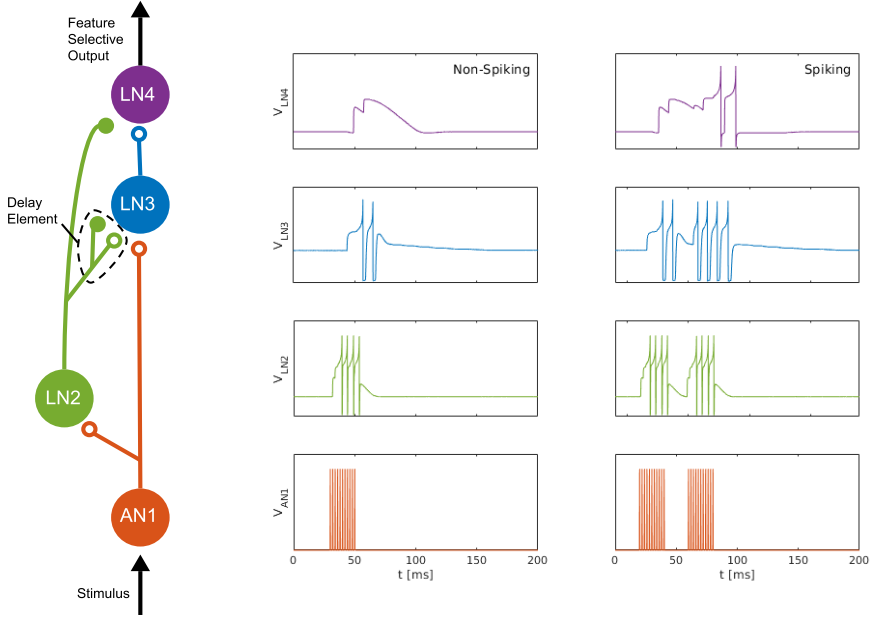}
    \end{center}
    \caption
    {
        Neuromorphic feature detection circuit inspired by an auditory feature detection circuit in field crickets.
        \textbf{(A)}~\ac{SNN} architecture comprising four spiking neurons, on which open circles and solid disks denote, respectively, excitatory and inhibitory synapses.
        The synaptic delay element imitates the dynamics of the non-spiking delay neuron LN5 in the feature detection circuit of the cricket \citep{schoneich2015auditory}. 
        \textbf{(B)}~Measured neuron membrane potentials in the DYNAP-SE, following a 20-ms pulse stimulus.
        \textbf{(C)}~Similarly, membrane potentials resulting from a pair of 20-ms stimuli pulses with a 20-ms interval which cause the feature detecting neuron LN4 to fire.
}
    \label{fig:featDet_network}
\end{figure}

We consider the auditory feature detection circuit for sound pattern recognition in the brain of female field crickets described by \cite{schoneich2015auditory}.
The circuit consists of five neurons and is used for the recognition of the sound pulse pattern of the male calling song, and it relies on a detection mechanism that selectively responds to the coincidence of a direct neural response and a delayed response to the received sound pulses.
In this circuit, a coincidence detecting neuron LN3 receives excitatory projections along two separate pathways;
one directly from the ascending neuron AN1,
and the other via the inhibitory neuron LN2 followed by a non-spiking delay neuron LN5,
which we approximate here with a delay element formed by an inhibitory-excitatory synapse pair, see Figure~\ref{fig:featDet_network} (adapted from \cite{nilsson2018monte}).
The non-spiking inhibitory neuron LN5 in the cricket projects to LN3 and provides a delayed excitation of LN3 due to \ac{PIR}.
The duration of the delay matches that of the species-specific sound \ac{IPI} that the circuit is specialized for detecting, so that the delayed excitation arrives at the coincidence detecting neuron LN3 simultaneously with the excitation caused by the subsequent sound pulse.
The coincident excitations of LN3 enables it to fire and excite the feature detecting neuron LN4.

\subsection{Disynaptic Delay Elements}

%
The \ac{PIR} of the non-spiking neuron LN5 in the cricket auditory feature detection circuit provides the delayed excitation of LN3 required for feature detection.
For a general discussion of such delays, see \cite{Buonomano1129} and \cite{mauk2004temporal}.
Spike based dynamic neuromorphic processors, such as the DYNAP-SE, cannot directly implement non-spiking neurons, such as the LN5 neuron in the cricket circuit, and flexible routing of such analog signals is problematic.
Therefore, we approximate LN5 and \ac{PIR} with an inhibitory--excitatory pair of dynamic synapses with different time constants, so that the sum of the two postsynaptic currents initially is inhibitory and subsequently becomes excitatory some time after presynaptic stimulation.
For the inhibitory effect, a synapse of the subtractive type is used in the DYNAP-SE.
As its name implies, the subtractive inhibitory synapse type allows for combining excitation and inhibition dynamics by summing inhibitory and excitatory postsynaptic currents, as opposed to the shunting synapse type which inhibits the neuron using a different mechanism.
This summation of postsynaptic currents is the central mechanism of the the proposed synaptic delay element.
For the excitatory part, the slow synapse type is used, leaving the fast synapse type available for bias configuration and use for stimulation of the postsynaptic neuron; in this case for the projection from AN1 to LN3.

\begin{figure}[h!]
    \begin{center}
        \includegraphics[width=0.8\textwidth]{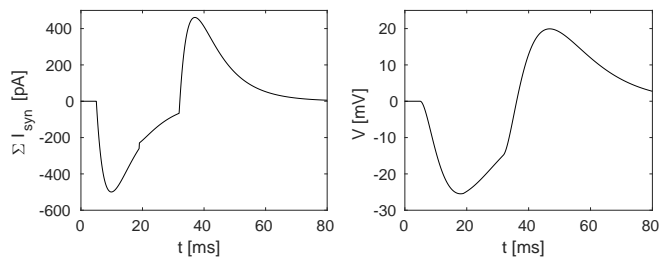}
    \end{center}
    \caption
    {
        Simulation of the synaptic delay element model.
        \textbf{(A)} Sum of inhibitory and excitatory postsynaptic currents from the delay element.
        \textbf{(B)} Resulting postsynaptic neuron membrane potential.
    }
    \label{fig:delay_sim}
\end{figure}

The proposed synaptic delay element can be modelled with Equation~\ref{eq:DPI},
and the membrane potential resulting from presynaptic stimulation can be illustrated by solving Equation~\ref{eq:AdEx}.
Figure~\ref{fig:delay_sim} shows a numerical simulation of the synaptic delay element model for a 20 ms constant input current that represents the presynaptic stimulation, as in Figure~\ref{fig:featDet_network}.
Since the simulated neuron is in the subthreshold regime ($V \ll V_T$),
Equation \ref{eq:AdEx} is simplified by setting the exponential term to zero and omitting the adaption variable.
The neuron and synapse parameters are selected so that the membrane potential is comparable to the potential measured in the hardware, and should thus not be directly compared with biological potentials and threshold values.

Dynamic synaptic elements of this type are expected to provide a delayed excitation that qualitatively matches the effect of \ac{PIR} in the output of non-spiking delay neurons like the LN5.
Furthermore, we expect that the time delay and relative amplitude of inhibition and excitation can be configured in a flexible way, for example by modifying the synapse time constants and efficacies.
The experimental results presented below demonstrate that this is indeed feasible,
and that for some bias settings it is possible to control the delay time and delayed excitation amplitude with the synaptic efficacies only.

\subsubsection{Hardware Configuration} 
\label{sec:hardwareImpl}

The synaptic delay elements were configured in the DYNAP-SE in the following way.
The delay elements were stimulated with four spikes equally spaced over the $\sim 20$~ms stimulus-response of LN2 for a 20-ms sound pulse, which represents the projection from LN2 to LN5 in the cricket circuit.
%
The time constant of the inhibitory synapse of the delay element was set so that the resulting inhibition of LN3 corresponds to the inhibition caused by LN5 in the cricket; that is, a couple of ms longer than the 20~ms sound pulse duration.
The excitatory synapse was tuned so that the LN3 excitation lasts somewhat longer than that of the initial inhibition,
approximately to the end of the corresponding \ac{PIR} excitation of LN5 in the cricket.
The weight of the inhibitory synapse was set higher than that of the excitatory synapse, such that the sum of inhibition and excitation turned out negative, thus inhibiting the neuron for the duration of the delay.
For the excitatory synapse, the weight was set to yield a substantial excitation of the postsynaptic neuron following the inhibition, while not generating spikes without additional synaptic stimulation.
In this manner, the effect of the non-spiking LN5 on LN3 is imitated with the summation of an inhibitory postsynaptic current and an excitatory postsynaptic current produced by the synapses of LN3 itself.

\subsubsection{Characterization}

The synaptic delay elements implemented in the DYNAP-SE were characterized by measuring the membrane potential of the postsynaptic neuron with an oscilloscope.
To avoid time synchronization issues, we analyzed the membrane potential measurements without reference to the precise timing of the presynaptic stimulation in terms of the full width at half minimum/maximum (FWHM) of the postsynaptic inhibition/excitation.
We characterized the synaptic delay elements with the distributions of the following five quantities:
the minimum membrane potential, $V_{min}$,
the maximum membrane potential, $V_{max}$,
the FWHM of inhibition, $\tau_{inh}$,
the FWHM of excitation, $\tau_{exc}$,
and the time duration from the FWHM onset of inhibition to the FWHM onset of excitation, $\tau_{delay}$.
These quantities are illustrated in Figure~\ref{fig:delay_char}, and allowed us to investigate the effect of different bias parameter settings on the synaptic delay elements in a population of neurons in the DYNAP-SE.
%
This way the bias parameter values of the delay elements could be tuned to imitate the behavior of the delay neuron LN5 in the cricket.
Further details on the experimental settings are described in Section \ref{sec:expSetting}.

\subsection{Neuromorphic Feature Detection Circuit}

For the implementation of the cricket auditory feature detection circuit in the DYNAP-SE neuromorphic processor, stimuli representing the projections from AN1 upon auditory stimulation were generated in the form of 11 spikes evenly distributed over the pulse duration of 20 ms (in the noise-free case), yielding 10 \acp{ISI} of 2 ms each.
Each of the remaining three neurons of the circuit, see Figure~\ref{fig:featDet_network}, were modeled on separate cores in one chip of the DYNAP-SE.

For the implementation of the inhibitory neuron LN2, a single neuron on a reserved core was used. This neuron was set to receive the generated stimulation representing AN1 by assigning a synaptic connection of the fast excitatory type. The bias parameter values from section 5.7.3 in the DYNAP-SE user guide \footnote{https://aictx.ai/technology/} were used as reference. The parameter values of the fast excitatory synapse were then adjusted in order to model the behavior of LN2 as observed in the cricket. The synaptic time constant \texttt{NPDPIE\_TAU\_F\_P} was adjusted to match that of the cricket, and the synapse weight \texttt{PS\_WEIGHT\_EXC\_F\_N} and threshold parameter \texttt{NPDPIE\_THR\_F\_P} were adjusted for LN2 to respond with the right amount of four to five spikes for each input pulse.

For the coincidence detecting neuron LN3, the proposed delay elements were implemented according to the earlier description. An excitatory connection of the fast type was added for LN3 to receive the projection from AN1.

For the excitatory connection from LN3 to the feature detecting neuron LN4, a synapse of the fast type was used, and, for the inhibitory connection from LN2 to LN4, a synapse of the subtractive type was used. Bias parameter values from section 5.7.3 in the DYNAP-SE user guide were used for neuronal parameters, and as reference values for the fast excitatory synapses. For the fast inhibitory synapse, bias values from section 5.7.5 in the user guide were used as reference. The bias parameters, time constant, threshold and weight, for both synapse types, were then hand-tuned in order to approximate the behavior of LN4 in one DYNAP-SE neuron.

\subsection{Experiments}
\label{sec:expSetting}

In all of the experiments conducted in this work, the DYNAP-SE neuromorphic processor was controlled using the cAER event-based processing framework for neuromorphic devices. More specifically, a custom module making use of the tools for configuration and monitoring provided by cAER was created and added to the framework. All stimuli were synthetically generated using the built-in spike generator in the \ac{FPGA} of the DYNAP-SE, which generates spike-events according to assigned \acp{ISI} and virtual source-neuron addresses.

The DYNAP-SE features analog ports for monitoring of neuron membrane potentials. For measurements of these potentials, the 8-bit USB oscilloscope SmartScope by LabNation was used. Since these measurements only capture the neuron membrane potential, there is no information about the precise relative timing of spike-events in the resulting data. Because of this, the durations of inhibition and excitation of the delay elements were defined in terms of the FWHM as described above.

For the extraction of the delay parameters defined in Section \ref{sec:hardwareImpl}, the stimulus was repeatedly broadcast to all neurons in the core, and for each stimulation cycle one neuron was monitored with the oscilloscope using the programmable analog outputs of the DYNAP-SE. The stimulation cycle was given a duration of 0.5~s, in order for the neurons to relax to a resting state before and after stimulation. At the initial state of rest, the resting potential was automatically estimated for each neuron. The resting potential was subsequently subtracted from the measurement data, such that the resulting resting potentials are zero.
This was done to make the parameter values of the different neurons comparable with each other.

%% file: Sections/3_results.tex
\subsection{Characterization of Delay Elements}

\begin{figure}[h!]
    \begin{center}
        \includegraphics[width=0.8\textwidth]{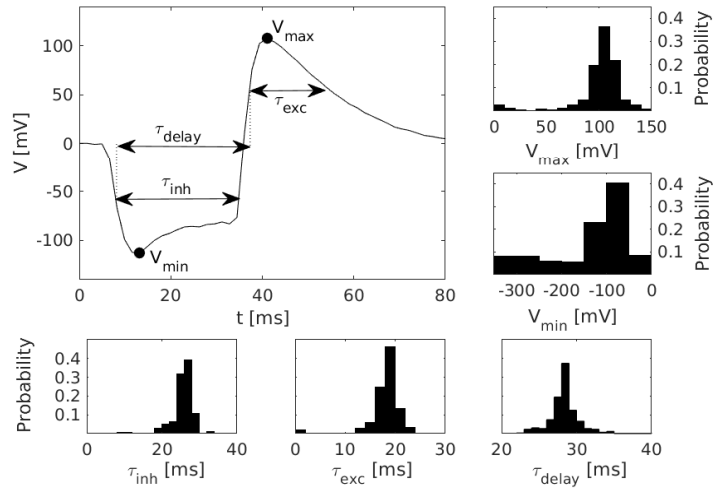}
    \end{center}
    \caption
    {
        Characterization of synaptic delay elements configured in the DYNAP-SE neuromorphic processor.
        \textbf{(A)} Postsynaptic membrane potential versus time, illustrating the delayed excitation resulting from a presynaptic pulse.
        \textbf{(B)} Distribution of the maximum measured membrane potential, $V_{max}$, resulting from a presynaptic pulse.
        \textbf{(C)} Similarly, the distribution of the minimum measured membrane potential, $V_{min}$.
        \textbf{(D)} Distribution of the inhibitory timescale, $\tau_{inh}$, defined as the full width at half minimum.
        \textbf{(E)} Distribution of the excitatory timescale, $\tau_{exc}$, defined as the full width at half maximum.
        \textbf{(F)} Distribution of the delay time, $\tau_{delay}$, defined as the duration from the onset of $\tau_{inh}$ to the offset of $\tau_{exc}$.
        The distributions in panels \textbf{(B)}-\textbf{(F)} are obtained via characterization of one DYNAP-SE core comprising 256 neurons with biases configured according to Table~\ref{tab:biases_delayElement}.
    }
    \label{fig:delay_char}
\end{figure}

The delay elements were implemented in one core of a DYNAP-SE neuromorphic processor; one delay element on each of the 256 neurons in the core. Results from the characterization of the delay elements are presented in Figure~\ref{fig:delay_char}.
The figure shows the pulse-response of one typical delay element from the resulting population, along with histograms of the distribution of parameters that characterize each delay element.
The resulting values of $V_{max}$ range from 3 to 143~mV and center around 105~mV.
$V_{min}$ have a thicker tail of the distribution and range from -310 to -20~mV, with most values between -100 and -50~mV.
The time constant distributions have relatively thin tails.
$\tau_{inh}$ has values between 6 and 47~ms with probability peaking between 26 and 28 ms.
$\tau_{exc}$ ranges from 0 to 38~ms with probability peaking between 18 and 20~ms, and $\tau_{delay}$ spans between 22 and 51~ms with probability peaking between 28 and 29~ms.

\begin{figure}[h!]
    \begin{center}
        \includegraphics[width=0.8\textwidth]{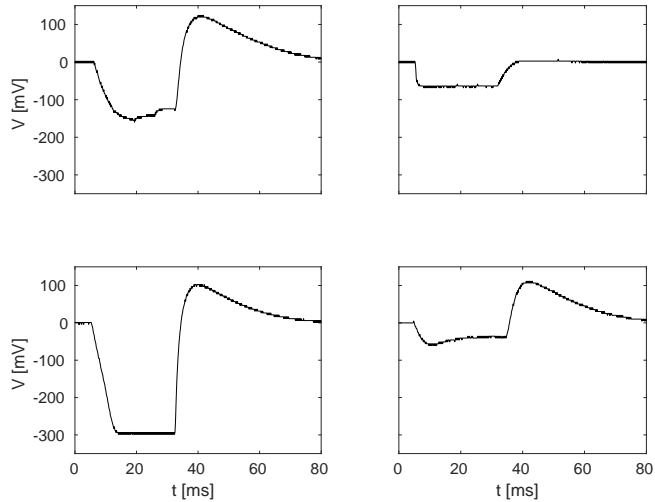}
    \end{center}
    \caption
    {
        Examples of four different membrane potentials measured in the characterization of delay elements summarized in Figure~\ref{fig:delay_char}.
        These variations are observed in one core with 256 neurons with biases configured according to Table~\ref{tab:biases_delayElement}.
    }
    \label{fig:delay_waveForms}
\end{figure}

The pulse-responses of four different delay elements are presented in Figure~\ref{fig:delay_waveForms}, which illustrates the variety of delay dynamics obtained thanks to device mismatch.
Here, the variance of the minimum voltage, $V_{min}$, is especially evident, but variation in other parameters can also be observed, such as $V_{max}$, in the case of the virtually non-existing excitation in Figure~\ref{fig:delay_waveForms}B.

\subsection{Cricket Feature Detector}
\label{sec:crickFeatDet}

\begin{figure}[h!]
    \begin{center}
        \includegraphics[width=0.8\textwidth]{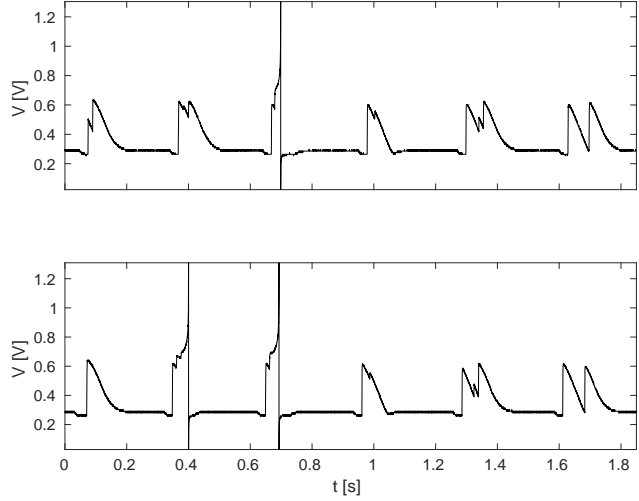}
    \end{center}
    \caption
    {
        Response of LN4 for double-pulse stimuli with \acp{IPI} of 0, 10, 20, 30, 40, and 50 ms, respectively.
        \textbf{(A)} Noiseless case.
        \textbf{(B)} Example for 20\% noise, with a false positive for the 10-ms \ac{IPI}.
    }
    \label{fig:featDet_LN4}
\end{figure}

The function of the feature detection network was investigated by stimulating it with double pulses of 20 ms duration each, while increasing the \ac{IPI} from 0, 10, 20, 30, 40, to 50 ms. Furthermore, in order to investigate the effect of noise in the stimuli, as is likely to be present in real-world environments, different levels of phase noise was introduced in the generated stimuli by randomly perturbing the value of the \acp{ISI} with values drawn from a continuous uniform distribution. Figure~\ref{fig:featDet_LN4} shows the membrane potential of LN4 during correct classification of noiseless double pulses of all of the \acp{IPI} mentioned above, as well as the result in the presence of 20\% phase noise, where some false positives are observed for the 10 ms \ac{IPI}.

\begin{figure}[h!]
    \begin{center}
        \includegraphics[width=0.8\textwidth]{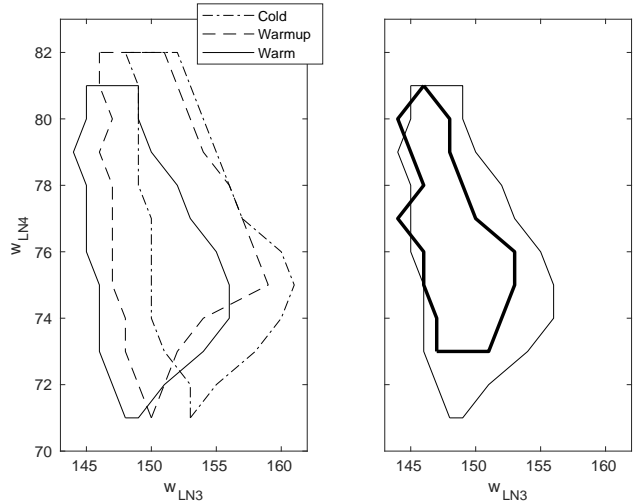}
    \end{center}
    \caption
    {
        Boundary of correct stimuli classification in synaptic parameter space. Outside the enclosed region, false positives and/or false negatives occur with varying probability. The horizontal (vertical) axis indicates the fine integer bias-value of the LN3 (LN4) excitatory synapse weight. Multiple line types indicate experiments performed under different environmental conditions.
        \textbf{(A)} Movement of the classification boundary observed after several hours of continuous operation from cold startup.
        The temperature change is likely caused by the FPGA that is enclosed in the system.
        \textbf{(B)} Shrinkage of the classification boundary in presence of 10\% phase noise in the stimuli (bold line). Boundary points are temperature dependent.
    }
    \label{fig:weightSpace}
\end{figure}

By varying the weights of the excitatory projection from AN1 to LN3 and the excitatory synapse weight of LN4, respectively, a boundary of correct classification of stimuli could identified in the space spanned by these two parameters.
Outside the boundary, false positives and/or false negatives occur with varying probability.
The boundary was observed to move substantially in the parameter space as time progressed after cold startup of the DYNAP-SE and this is likely due to heating by the FPGA that is enclosed in the DYNAP-SE system.
This change was observed over multiple runs of the experiment and appears to be qualitatively consistent.
Furthermore, the shift of the boundary in the presence of phase noise in the stimuli was investigated. Figure~\ref{fig:weightSpace} shows the boundary of correct classification, as measured at three separate points in time after device initialization, spanning from minutes to several hours of run-time. The figure also shows the shrinkage of the classification boundary in the presence of 10\% phase noise in the stimuli, in relation to the steady-state of the boundary after several hours of system run-time.

\begin{figure}[h!]
    \begin{center}
        \includegraphics[width=0.8\textwidth]{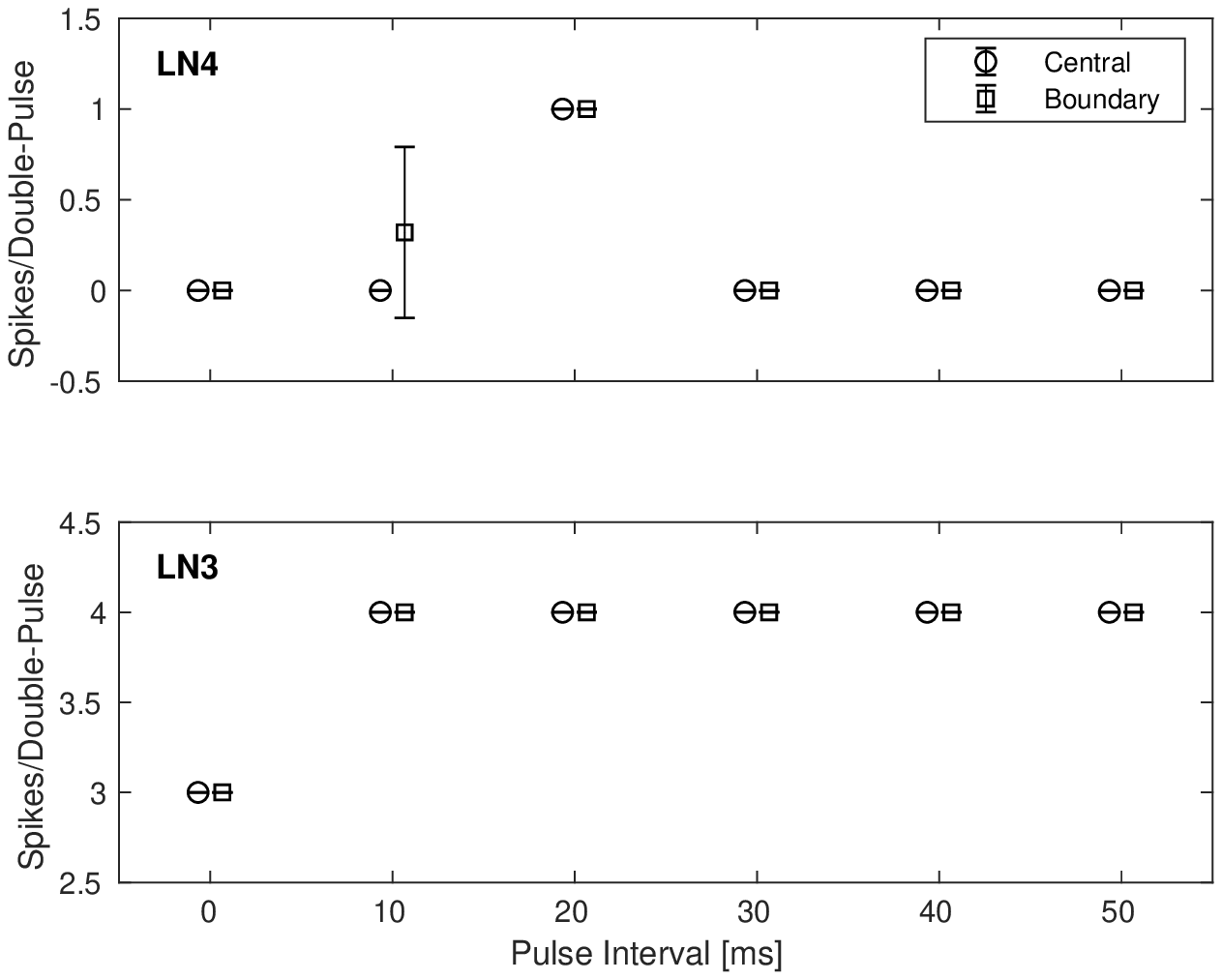}
    \end{center}
    \caption
    {
        Average number of spikes from LN3 and LN4 per double-pulse stimuli for varying \acp{IPI} and two different bias configurations, centrally and on the boundary illustrated in Figure~\ref{fig:weightSpace}.
        For each \ac{IPI}, the data-points are graphically separated by 4/3 ms to improve clarity of the visualization.
        Error bars denote $\pm 1$ standard deviation.
        \textbf{(A)} Feature detecting neuron, LN4.
        \textbf{(B)} Coincidence detecting neuron, LN3.
    }
    \label{fig:spikeCount_weightSpace}
\end{figure}
A quantitative investigation of the \ac{IPI} dependence of the feature detection circuit was made by repeatedly stimulating the network with double pulses of different \acp{IPI} as described earlier, while observing the response in LN3 and LN4 by recording and counting the spikes of both neurons.
For each \ac{IPI}, the network was presented with the corresponding double-pulse stimulus 50 times. Figure~\ref{fig:spikeCount_weightSpace} shows, in the case of noiseless stimuli, the average number of spikes from LN3 and LN4, respectively, centrally within the synaptic boundary of correct classification, as well as at the boundary. Centrally within the boundary of correct classification, LN4 responded exclusively to the 20 ms pulse interval, with no false positives or negatives.
On the boundary of the parameter space, LN4 began to exhibit false positives for the 10 ms \ac{IPI}, with 0.32 $\pm$ 0.47 spikes per double-pulse stimulus.

\begin{figure}[h!]
    \begin{center}
        \includegraphics[width=0.8\textwidth]{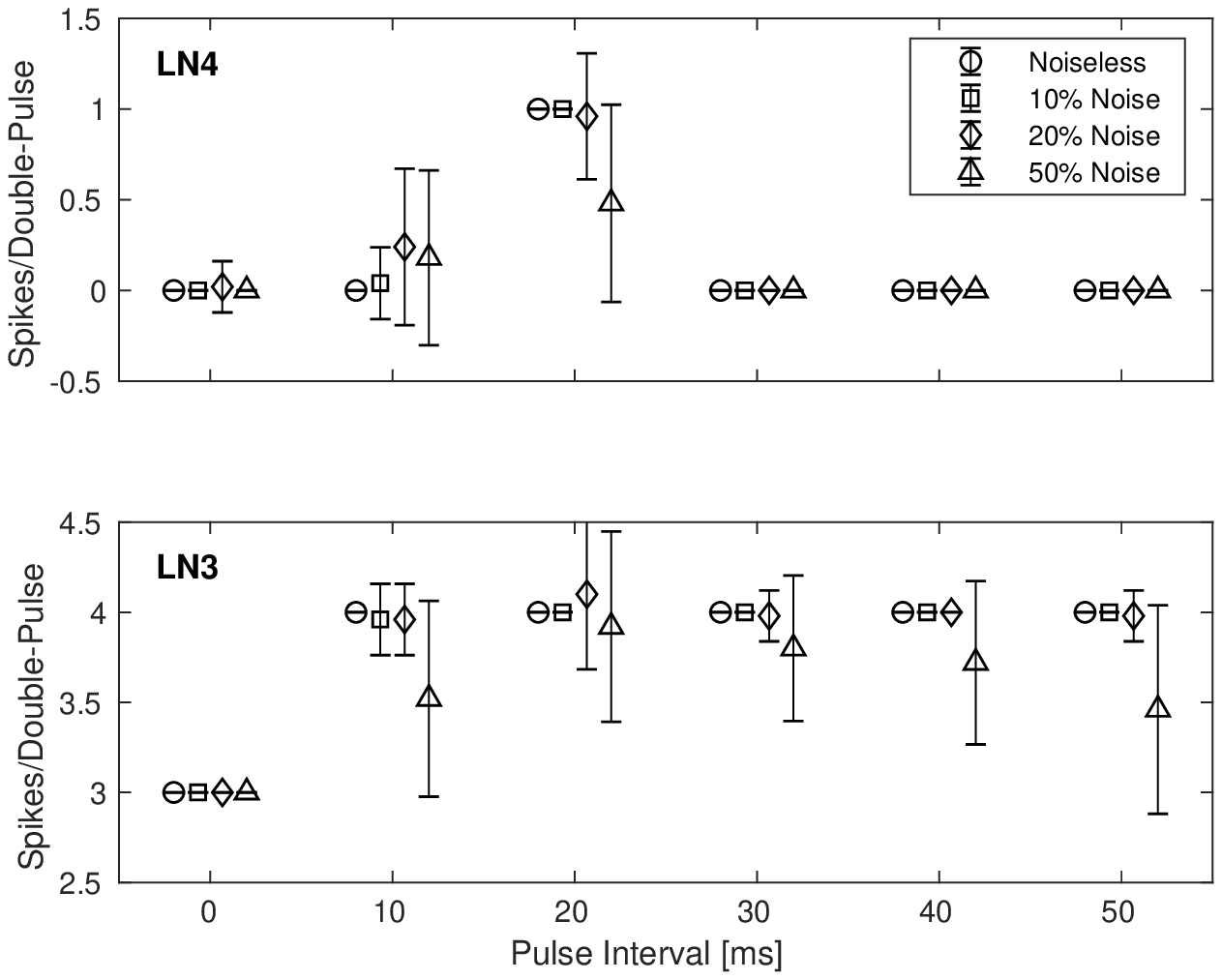}
    \end{center}
    \caption
    {
        Average number of spikes from LN3 and LN4 per double-pulse stimulus for varying \acp{IPI} and different levels of phase noise in the stimuli.
        For each \ac{IPI}, the data-points are graphically separated by 4/3 ms to improve clarity of the visualization.
        Error bars denote $\pm 1$ standard deviation.
        \textbf{(A)} Feature detecting neuron, LN4.
        \textbf{(B)} Coincidence detecting neuron, LN3.
    }
    \label{fig:spikeCount_noise}
\end{figure}

Similarly, Figure~\ref{fig:spikeCount_noise} shows the results for the best synaptic configuration used in the previous experiment, centrally located within the boundary of correct classification, but for different levels of phase noise.
As expected the network performed correct classification in the noiseless case.
The introduction of noise caused LN4 to exhibit false positives,
in particular for the 10 ms \ac{IPI}.
At higher levels of noise also false negatives were observed.
In the case of 50\% noise the response of LN4 was 0.18 $\pm$ 0.48 spikes per double-pulse for the 10 ms \ac{IPI}, and 0.48 $\pm$ 0.54 spikes for the 20 ms \ac{IPI}.

\subsection{Configuration of Delay Elements}

\begin{figure}[h!]
    \begin{center}
        \includegraphics[width=0.8\textwidth]{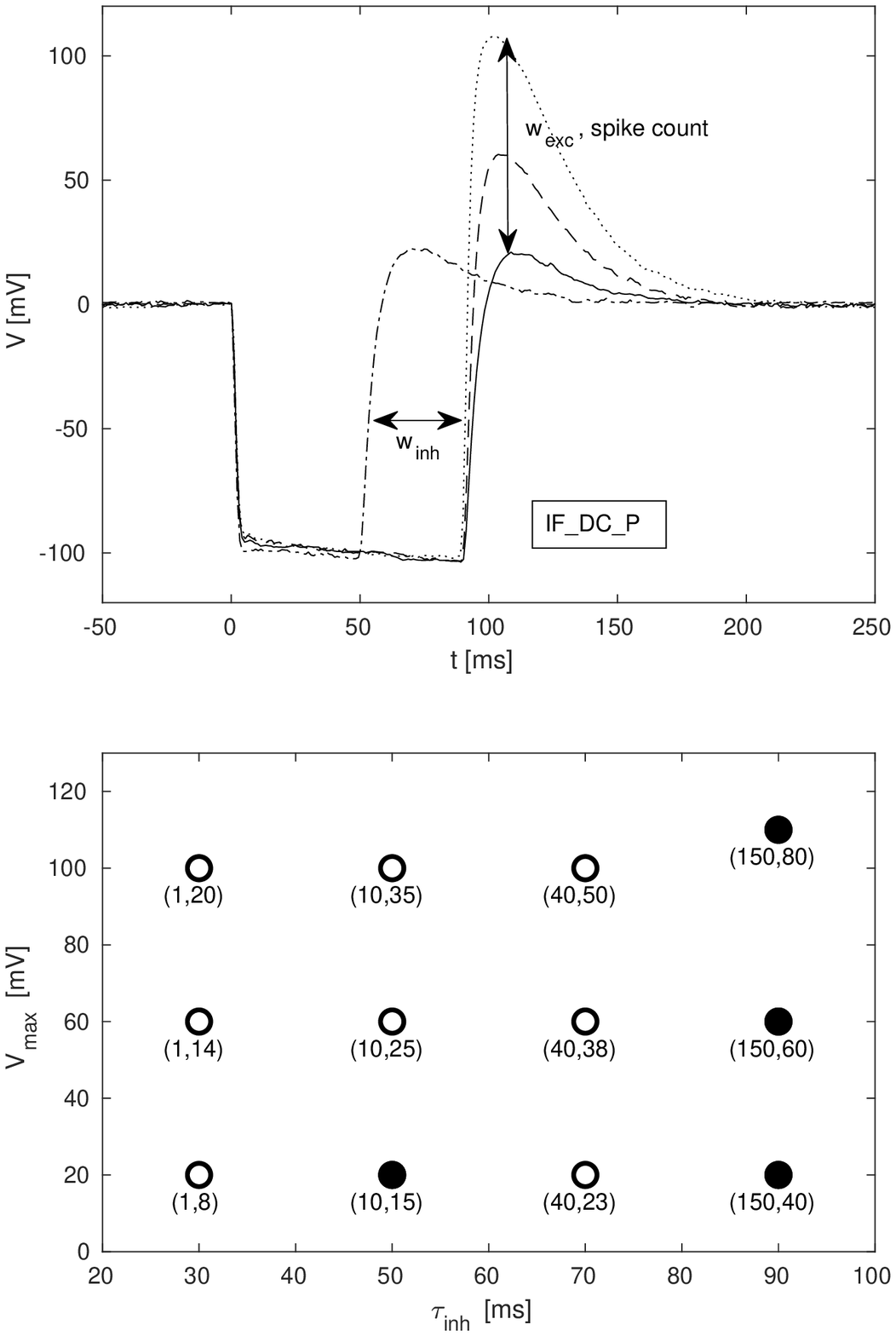}
    \end{center}
    \caption
    {
        Configuration of synaptic delay elements.
        \textbf{(A)} Postsynaptic membrane potential versus time resulting from a presynaptic pulse.
        The delay is controlled mainly by the inhibitory synaptic efficacy, $w_{inh}$.
        The amplitude of the delayed excitation is controlled mainly by the excitatory synaptic efficacy, $w_{exc}$, and the number of presynaptic spikes.
        Note that the membrane potential reaches its minimum possible value during inhibition, and that the difference between this value and the resting potential is controlled with the constant injection current of the neuron.
        \textbf{(B)} Maximum membrane potential versus duration of inhibition for different values of $(w_{inh}, w_{exc})$.
        Each point is denoted with the corresponding fine integer bias-values of the inhibitory and excitatory synapse weights, respectively.
    }
    \label{fig:delay_variations}
\end{figure}

Given the large parameter space of a dynamic neuromorphic processor like the DYNAP-SE we explored different ways to simplify the configuration of the disynaptic delay elements for delays up to about 100~ms.
One possibility is to lower the constant injection current of the neurons receiving the delayed signal to such an extent that the inhibition by the delay elements make the neuron reach its minimum membrane potential.
This results in delay elements for which the duration of inhibition, $\tau_{inh}$, can be controlled with the inhibitory weight of the delay element, $w_{inh}$.
In this case the amplitude of the postinhibitory excitation, $V_{max}$, is controlled by the excitatory weight of the delay element, $w_{exc}$, as well as by varying the number of incoming spikes stimulating the delay element. Figure~\ref{fig:delay_variations}A shows four configurations of one delay element, with the maximum membrane potential of the postinhibitory excitation ranging between 20 and 110~mV, and the durations of inhibition ranging between 50 and 90~ms, according to the \ac{FDHM} definition.
A table with delay element weight values and resulting values of $\tau_{inh}$ and $V_{max}$, from a total of 12 such variations, is presented in Figure~\ref{fig:delay_variations}B; the data-points corresponding to the membrane potentials in Figure~\ref{fig:delay_variations}A are marked with filled disks.

%% file: Sections/4_discussion.tex
%

Temporal feature detection and pattern recognition are central tasks in advanced sensor and perception systems.
Thus, low-power solutions enabling learning and recognition of complex patterns with less energy has many potential applications, for example in embedded intelligence and deep edge sensor systems. 
In particular, ultra-low power solutions operating at the order of milliwatts is a key enabler for advanced wireless sensor systems,
for example for machine monitoring \citep{Campo2017cm,Campo2013ml} where the system needs to operate autonomously with limited resources over the expected lifetime of the monitored machine \citep{Haggstrom2018thesis,Campo2017thesis}.

Searching for effective \ac{SNN} architectures for pattern recognition that are suitable for implementation in ultra-low power dynamic neuromorphic processors like the DYNAP-SE, we adapted and investigated the aforementioned cricket auditory feature detection circuit.
Surprisingly, we found that the conceived delay elements formed by two dynamic synapses can be easily configured in terms of the desired delay and excitation amplitude by changing only the synaptic efficacies, and that the \ac{PIR}-based delay in the cricket circuit can be qualitatively reproduced in this way.
Although we sidestep Dale's principle, the resulting dynamic synaptic delay elements have the desirable property that a single neuron with high fan-in can integrate multiple temporal features.
Furthermore, by adjusting the bias settings of the synapses, long delays of order 10-100 ms are efficiently realised, and, since the delay element requires only standard dynamic synapses, it can be implemented in other dynamic neuromorphic processors where synaptic plasticity can enable tuning of temporal feature detectors.

At the quantitative level, we observe some differences between the feature detection results presented in Section~\ref{sec:crickFeatDet} and the behaviour of the cricket circuit described by \cite{schoneich2015auditory}.
In the crickets, the response of the coincidence detector neuron LN3 for different \acp{IPI} varies so that the distribution of the number of spikes of LN3 increases as the interval gets closer to the species-specific \ac{IPI} of 20~ms.
This is not the case in the results presented here, and further optimization of the neuron and synapse parameters are required if this behaviour is to be imitated.
As illustrated in Figure \ref{fig:spikeCount_weightSpace}B, our LN3 reliably produces the same number (but different timings) of spikes for all of the different \acp{IPI}, with the exception of the 0~ms \ac{IPI}.
A more plausible trend is observed in the case of $50 \%$ input noise, but in that case the classification results are less encouraging.
%
Hence, the classification mechanism relies on the timing of spikes and the balance of inhibition and excitation.

\begin{figure}[h!]
    \begin{center}
        \includegraphics[width=0.6\textwidth]{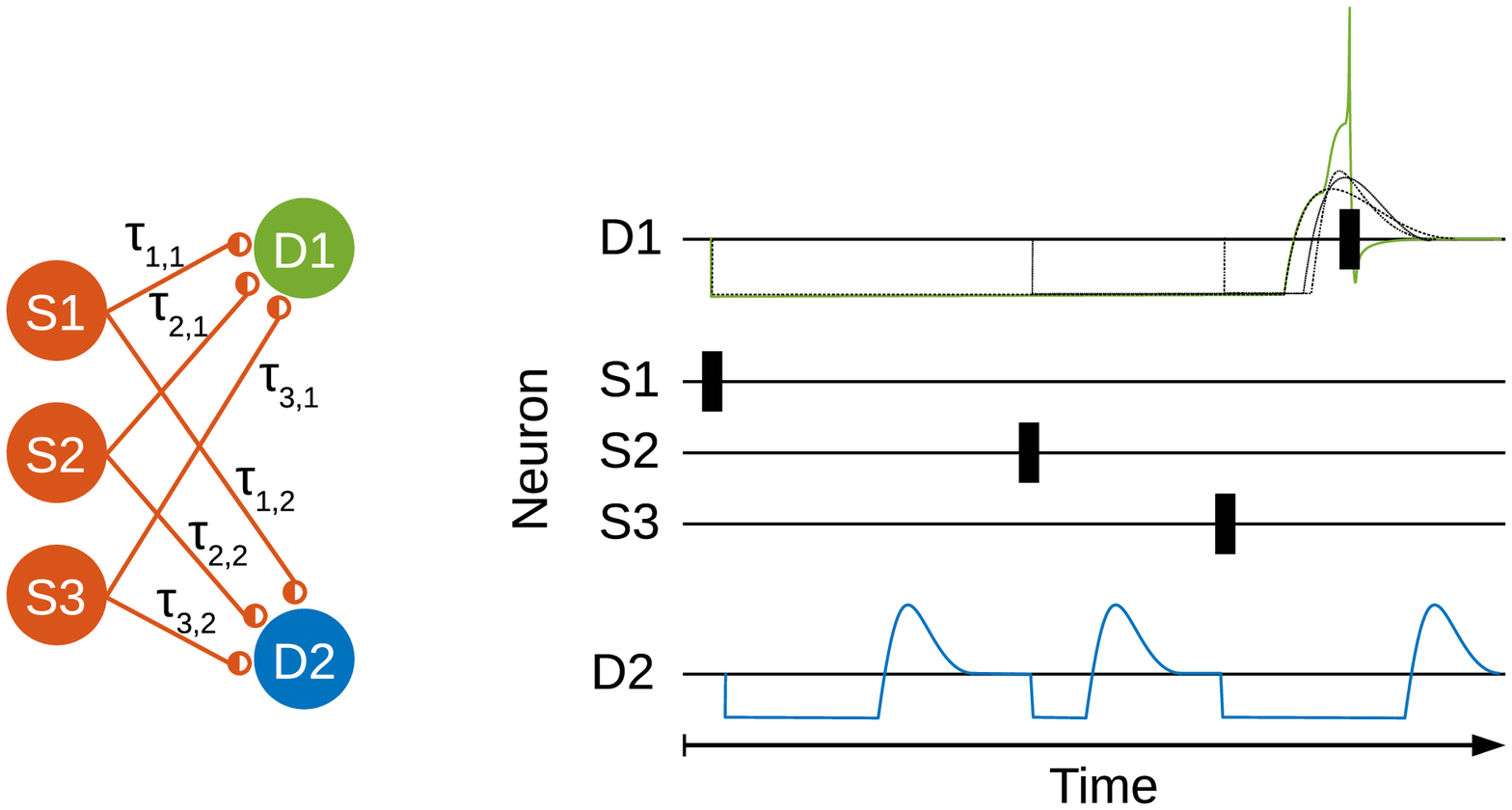}
    \end{center}
    \caption
    {
        Spatiotemporal pattern recognition in a polychronous network.
        \textbf{(A)} Source neurons $S1$, $S2$ and $S3$ project to detector neurons $D1$ and $D2$ via synaptic delay elements (semi-filled disks) with different delays $\tau_{i,j}$.
        \textbf{(B)} A spatiotemporal pattern of spikes (vertical bars) emitted from $S1-S3$ is detected by $D1$ due to temporally coinciding delayed excitatory postsynaptic currents, which raise the membrane potential of $D1$ (green line) above the spiking threshold.
        The membrane potential of $D2$ (blue line) remains below threshold because the excitatory postsynaptic currents peak at different points in time.}
        \label{fig:polyNet}
\end{figure}

By combining multiple synaptic delay elements as illustrated in Figure \ref{fig:polyNet}, for example in line with the idea of polychronous networks \citep{izhikevich2006polychronization}, arbitrary spatiotemporal patterns can be detected.
Further work is required to investigate how the repertoire of synaptic delays in one core of the DYNAP-SE should be exploited and configured/learned to solve practical pattern recognition tasks, and to further develop the understanding of how device mismatch, noise and  temperature variations affect different network architectures.
%
%
%
%
%
%
%
%
%
%
%
In general, with dynamic synapses featuring short- and long-term plasticity, more sophisticated mechanisms for sequence detection and learning can also be realised \citep{Buonomano1129} and investigated.
The NMDA controlled excitatory synapses exhibit a rich repertoire of dynamic behavior which can be further explored for synaptic delay and pattern recognition purposes.
Furthermore, \acp{SNN} can faithfully reproduce the dynamics of brain networks, which appears to self-organize near a critical point where no privileged spatial or temporal scale exist, which has interesting consequences for information processes \citep{Cocchi2017criticality}.
Thus, Neuromorphic Engineering based on dynamic neuromorphic processors opens the way to new interesting architectures for pattern recognition and ultra low-power adaptive solutions to pattern recognition and generation tasks in machine perception, learning and control.
%


%% file: backmatter.tex
\section*{Conflict of Interest Statement}

The authors declare that the research was conducted in the absence of any commercial or financial relationships that could be construed as a potential conflict of interest.

\section*{Author Contributions}

FS conceived the possibility to imitate non-spiking PIR delay in DYNAP-SE with synaptic dynamics, supervised the experiments to be carried out, and wrote part of the manuscript.
MN implemented the code that controls the DYNAP-SE, performed the experiments, and wrote part of the manuscript.

\section*{Funding}
This work is supported by The Kempe Foundations under contract JCK-1809 and SMK-1429, and was enabled by a collaboration with the Institute of Neuroinformatics in Zurich supported by STINT under contract IG2011-2025.

\section*{Acknowledgments}
Ideas leading to the work presented here have been discussed at the CapoCaccia Neuromorphic Engineering Workshop, in particular with Giacomo Indiveri, and the bias parameters of the delay elements presented in Table~\ref{tab:biases_delayElement} are based on biases kindly shared by Nicoletta Risi.
We thank Federico Corradi and Carsten Nielsen for technical support with the DYNAP-SE neuromorphic system.
This work would not have been possible without long-term strategic support from our university.
In particular, we thank Jerker~Delsing for supporting the developments in Neuromorphic Engineering at EISLAB and Jonas Ekman for support at the departmental level.



%% file: Sections/A_bias_values.tex

\begin{table}[h!]
    \caption
    {
        Bias parameter values used for the characterization of individual disynaptic delay elements in the DYNAP-SE.
    }
    \vspace{0.5cm}
    \centering
	\begin{tabular}{|l|l|c|c|c|}
	    \hline
	    Parameter Type & Parameter Name & Coarse Value & Fine Value & Current Level \\
	    
	    \hline
	    \multirow{11}{*}{Neuronal}
	    
	    & \texttt{IF\_AHTAU\_N} 			& 	7   & 	35  &   L   \\
	    & \texttt{IF\_AHTHR\_N} 			& 	7 	& 	1   &   H   \\
	    & \texttt{IF\_AHW\_P} 				&	7 	& 	1 	&   H   \\
	    & \texttt{IF\_BUF\_P}				& 	3	& 	80	&   H   \\
	    & \texttt{IF\_CASC\_N}				& 	7	& 	1	&   H   \\
	    & \texttt{IF\_DC\_P}				&	0	&	40	&   H   \\
	    & \texttt{IF\_NMDA\_N}				& 	1	& 	213	&   H   \\
   		& \texttt{IF\_RFR\_N}				&	4	& 	40	&   H   \\
   		& \texttt{IF\_TAU1\_N}				& 	5	& 	39	&   L   \\
   		& \texttt{IF\_TAU2\_N}				& 	0	&	15	&   H   \\
   		& \texttt{IF\_THR\_N}				&	6	& 	4	&   H   \\
   		
   		\hline
		\multirow{8}{*}{Synaptic}
		
   		& \texttt{NPDPIE\_TAU\_S\_P}			& 	6	& 	120	&   H   \\
   		& \texttt{NPDPIE\_THR\_S\_P}			& 	1 	&	30	&   H   \\
   		
   		& \texttt{NPDPII\_TAU\_F\_P}			& 	5	& 	100	&   H   \\
   		& \texttt{NPDPII\_THR\_F\_P}			& 	3	& 	60	&   H   \\
   		
   		& \texttt{PS\_WEIGHT\_EXC\_S\_N}		& 	1	& 	110	&   H   \\
   		& \texttt{PS\_WEIGHT\_INH\_F\_N}		& 	1	& 	130  &   H   \\
   		
   		& \texttt{PULSE\_PWLK\_P}				&	5	& 	40	&   H   \\
   		& \texttt{R2R\_P}						& 	4	& 	85	&   H   \\
   		\hline
	\end{tabular}
    \label{tab:biases_delayElement}
\end{table}

\begin{table}[h!]
    \caption
    {
        Bias parameter values used for the inhibitory neuron, LN2, in the DYNAP-SE implementation of the cricket feature detection network..
    }
    \vspace{0.5cm}
    \centering
	\begin{tabular}{|l|l|c|c|c|}
	    \hline
	    Parameter Type & Parameter Name & Coarse Value & Fine Value & Current Level \\
	    
	    \hline
	    \multirow{11}{*}{Neuronal}
	    
	    & \texttt{IF\_AHTAU\_N} 			& 	7   & 	35  &   L   \\
	    & \texttt{IF\_AHTHR\_N} 			& 	7 	& 	1   &   H   \\
	    & \texttt{IF\_AHW\_P} 				&	7 	& 	1 	&   H   \\
	    & \texttt{IF\_BUF\_P}				& 	3	& 	80	&   H   \\
	    & \texttt{IF\_CASC\_N}				& 	7	& 	1	&   H   \\
	    & \texttt{IF\_DC\_P}				&	7	&	2	&   H   \\
	    & \texttt{IF\_NMDA\_N}				& 	7	& 	1	&   H   \\
   		& \texttt{IF\_RFR\_N}				&	4	& 	208	&   H   \\
   		& \texttt{IF\_TAU1\_N}				& 	6	& 	21	&   L   \\
   		& \texttt{IF\_TAU2\_N}				& 	5	&	15	&   H   \\
   		& \texttt{IF\_THR\_N}				&	3	& 	20	&   H   \\
   		
   		\hline
		\multirow{5}{*}{Synaptic}
		
   		& \texttt{NPDPIE\_TAU\_F\_P}			&	5	&	165	&   H   \\
   		& \texttt{NPDPIE\_THR\_F\_P}			& 	1	& 	100	&   H   \\
   		
   		
   		& \texttt{PS\_WEIGHT\_EXC\_F\_N} 		&	0 	&	190	&   H   \\
   		
   		& \texttt{PULSE\_PWLK\_P}				&	0	& 	43	&   H   \\
   		& \texttt{R2R\_P}						& 	4	& 	85	&   H   \\
   		\hline
	\end{tabular}
    \label{tab:biases_LN2}
\end{table}

\begin{table}[h!]
    \caption
    {
        Bias parameter values used for the coincidence detecting neuron, LN3, in the DYNAP-SE implementation of the cricket feature detection network.
    }
    \vspace{0.5cm}
    \centering
	\begin{tabular}{|l|l|c|c|c|}
	    \hline
	    Parameter Type & Parameter Name & Coarse Value & Fine Value & Current Level \\
	    
	    \hline
	    \multirow{11}{*}{Neuronal}
	    
	    & \texttt{IF\_AHTAU\_N} 			& 	7   & 	35  &   L   \\
	    & \texttt{IF\_AHTHR\_N} 			& 	7 	& 	1   &   H   \\
	    & \texttt{IF\_AHW\_P} 				&	7 	& 	1 	&   H   \\
	    & \texttt{IF\_BUF\_P}				& 	3	& 	80	&   H   \\
	    & \texttt{IF\_CASC\_N}				& 	7	& 	1	&   H   \\
	    & \texttt{IF\_DC\_P}				&	0	&	40	&   H   \\
	    & \texttt{IF\_NMDA\_N}				& 	1	& 	213	&   H   \\
   		& \texttt{IF\_RFR\_N}				&	4	& 	40	&   H   \\
   		& \texttt{IF\_TAU1\_N}				& 	5	& 	39	&   L   \\
   		& \texttt{IF\_TAU2\_N}				& 	0	&	15	&   H   \\
   		& \texttt{IF\_THR\_N}				&	6	& 	4	&   H   \\
   		
   		\hline
		\multirow{11}{*}{Synaptic}
		
   		& \texttt{NPDPIE\_TAU\_F\_P}			&	5	&	200	&   H   \\
   		& \texttt{NPDPIE\_TAU\_S\_P}			& 	6	& 	120	&   H   \\
   		& \texttt{NPDPIE\_THR\_F\_P}			& 	1	& 	30	&   H   \\
   		& \texttt{NPDPIE\_THR\_S\_P}			& 	1 	&	30	&   H   \\
   		
   		& \texttt{NPDPII\_TAU\_F\_P}			& 	5	& 	100	&   H   \\
   		& \texttt{NPDPII\_THR\_F\_P}			& 	3	& 	60	&   H   \\
   		
   		& \texttt{PS\_WEIGHT\_EXC\_F\_N} 		&	1 	&	144--161	&   H   \\
   		& \texttt{PS\_WEIGHT\_EXC\_S\_N}		& 	1	& 	110	&   H   \\
   		& \texttt{PS\_WEIGHT\_INH\_F\_N}		& 	1	& 	130  &   H   \\
   		
   		& \texttt{PULSE\_PWLK\_P}				&	5	& 	40	&   H   \\
   		& \texttt{R2R\_P}						& 	4	& 	85	&   H   \\
   		\hline
	\end{tabular}
    \label{tab:biases_LN3}
\end{table}

\begin{table}[h!]
    \caption
    {
        Bias parameter values used for the feature detecting neuron, LN4, in the DYNAP-SE implementation of the cricket feature detection network.
    }
    \vspace{0.5cm}
    \centering
	\begin{tabular}{|l|l|c|c|c|}
	    \hline
	    Parameter Type & Parameter Name & Coarse Value & Fine Value & Current Level \\
	    
	    \hline
	    \multirow{11}{*}{Neuronal}
	    
	    & \texttt{IF\_AHTAU\_N} 			& 	7   & 	35  &   L   \\
	    & \texttt{IF\_AHTHR\_N} 			& 	7 	& 	1   &   H   \\
	    & \texttt{IF\_AHW\_P} 				&	7 	& 	1 	&   H   \\
	    & \texttt{IF\_BUF\_P}				& 	3	& 	80	&   H   \\
	    & \texttt{IF\_CASC\_N}				& 	7	& 	1	&   H   \\
	    & \texttt{IF\_DC\_P}				&	7	&	2	&   H   \\
	    & \texttt{IF\_NMDA\_N}				& 	7	& 	1	&   H   \\
   		& \texttt{IF\_RFR\_N}				&	4	& 	208	&   H   \\
   		& \texttt{IF\_TAU1\_N}				& 	6	& 	21	&   L   \\
   		& \texttt{IF\_TAU2\_N}				& 	5	&	15	&   H   \\
   		& \texttt{IF\_THR\_N}				&	3	& 	20	&   H   \\
   		
   		\hline
		\multirow{8}{*}{Synaptic}
		
   		& \texttt{NPDPIE\_TAU\_F\_P}			&	5	&	80	&   H   \\
   		& \texttt{NPDPIE\_THR\_F\_P}			& 	1	& 	140	&   H   \\
   		
   		& \texttt{NPDPII\_TAU\_F\_P}			& 	6	& 	180	&   H   \\
   		& \texttt{NPDPII\_THR\_F\_P}			& 	3	& 	140	&   H   \\
   		
   		& \texttt{PS\_WEIGHT\_EXC\_F\_N} 		&	0 	&	71--82	&   H   \\
   		& \texttt{PS\_WEIGHT\_INH\_F\_N}		& 	0	& 	60  &   H   \\
   		
   		& \texttt{PULSE\_PWLK\_P}				&	0	& 	43	&   H   \\
   		& \texttt{R2R\_P}						& 	4	& 	85	&   H   \\
   		\hline
	\end{tabular}
    \label{tab:biases_LN4}
\end{table}

\begin{table}[h!]
    \caption
    {
        Bias parameter values used for configuration of the disynaptic delay elements in the DYNAP-SE.
    }
    \vspace{0.5cm}
    \centering
	\begin{tabular}{|l|l|c|c|c|}
	    \hline
	    Parameter Type & Parameter Name & Coarse Value & Fine Value & Current Level \\
	    
	    \hline
	    \multirow{11}{*}{Neuronal}
	    
	    & \texttt{IF\_AHTAU\_N} 			& 	7   & 	35  &   L   \\
	    & \texttt{IF\_AHTHR\_N} 			& 	7 	& 	1   &   H   \\
	    & \texttt{IF\_AHW\_P} 				&	7 	& 	1 	&   H   \\
	    & \texttt{IF\_BUF\_P}				& 	3	& 	80	&   H   \\
	    & \texttt{IF\_CASC\_N}				& 	7	& 	1	&   H   \\
	    & \texttt{IF\_DC\_P}				&	1	&	30	&   H   \\
	    & \texttt{IF\_NMDA\_N}				& 	1	& 	213	&   H   \\
   		& \texttt{IF\_RFR\_N}				&	4	& 	40	&   H   \\
   		& \texttt{IF\_TAU1\_N}				& 	5	& 	39	&   L   \\
   		& \texttt{IF\_TAU2\_N}				& 	0	&	15	&   H   \\
   		& \texttt{IF\_THR\_N}				&	6	& 	4	&   H   \\
   		
   		\hline
		\multirow{8}{*}{Synaptic}
		
   		& \texttt{NPDPIE\_TAU\_S\_P}			& 	7	& 	210	&   H   \\
   		& \texttt{NPDPIE\_THR\_S\_P}			& 	1 	&	30	&   H   \\
   		
   		& \texttt{NPDPII\_TAU\_F\_P}			& 	6	& 	80	&   H   \\
   		& \texttt{NPDPII\_THR\_F\_P}			& 	3	& 	60	&   H   \\
   		
   		& \texttt{PS\_WEIGHT\_EXC\_S\_N}		& 	0	& 	8--80	&   H   \\
   		& \texttt{PS\_WEIGHT\_INH\_F\_N}		& 	0	& 	1--150  &   H   \\
   		
   		& \texttt{PULSE\_PWLK\_P}				&	5	& 	40	&   H   \\
   		& \texttt{R2R\_P}						& 	4	& 	85	&   H   \\
   		\hline
	\end{tabular}
    \label{tab:biases_delayVar}
\end{table}